\documentclass[runningheads]{llncs}
\usepackage{graphicx}
\usepackage{amsmath,amssymb} 
\usepackage{color}
\usepackage{xcolor}
\usepackage{multirow}
\usepackage{hyperref}
\usepackage{subfigure}
\usepackage{booktabs}
\hypersetup{colorlinks,linkcolor={red},citecolor={green},urlcolor={red}}
\begin{document}

\newcommand*\rfrac[2]{{}^{#1}\!/_{#2}}
\newcommand{\layername}[1]{{\fontfamily{qcr}\selectfont#1}}
\newcommand{\myparagraph}[1]{{\vspace{0.5em} \noindent \bf #1}}

\pagestyle{headings}
\mainmatter

\def\eg{\emph{e.g.}} \def\Eg{\emph{E.g.}}
\def\ie{\emph{i.e.}} \def\Ie{\emph{I.e}\onedot}
\def\cf{\emph{c.f.}} \def\Cf{\emph{C.f}\onedot}
\def\etc{\emph{etc}.} \def\vs{\emph{vs}\onedot}
\def\wrt{w.r.t.} \def\dof{d.o.f.}
\def\etal{\emph{et al.}}

\title{Unified Perceptual Parsing for Scene Understanding} 
%

\titlerunning{Unified Perceptual Parsing for Scene Understanding}

\authorrunning{T. Xiao, Y. Liu, B. Zhou, Y. Jiang, J. Sun}

\author{Tete Xiao\inst{1}*, Yingcheng Liu\inst{1}*, Bolei Zhou\inst{2}*, Yuning Jiang\inst{3}, Jian Sun\inst{4}}


\institute{\quad ${}^1$Peking University \quad ${}^2$ MIT CSAIL \quad ${}^3$ Bytedance Inc. \quad ${}^4$~Megvii Inc.\\ * indicates equal contribution. \\
\email{\{jasonhsiao97, liuyingcheng\}@pku.edu.cn,\\
bzhou@csail.mit.edu, jiangyuning@bytedance.com, \\
sunjian@megvii.com}
}

\maketitle

\begin{abstract}

Humans recognize the visual world at multiple levels: we effortlessly categorize scenes and detect objects inside, while also identifying the textures and surfaces of the objects along with their different compositional parts. In this paper, we study a new task called Unified Perceptual Parsing, which requires the machine vision systems to recognize as many visual concepts as possible from a given image. A multi-task framework called UPerNet and a training strategy are developed to learn from heterogeneous image annotations. We benchmark our framework on Unified Perceptual Parsing and show that it is able to effectively segment a wide range of concepts from images. The trained networks are further applied to discover visual knowledge in natural scenes\footnote{Models are available at \url{https://github.com/CSAILVision/unifiedparsing}}.

\keywords{Deep neural network, semantic segmentation, scene understanding}
\end{abstract}

\section{Introduction}

\begin{figure}[!t]
\centering
\includegraphics[width=1\linewidth]{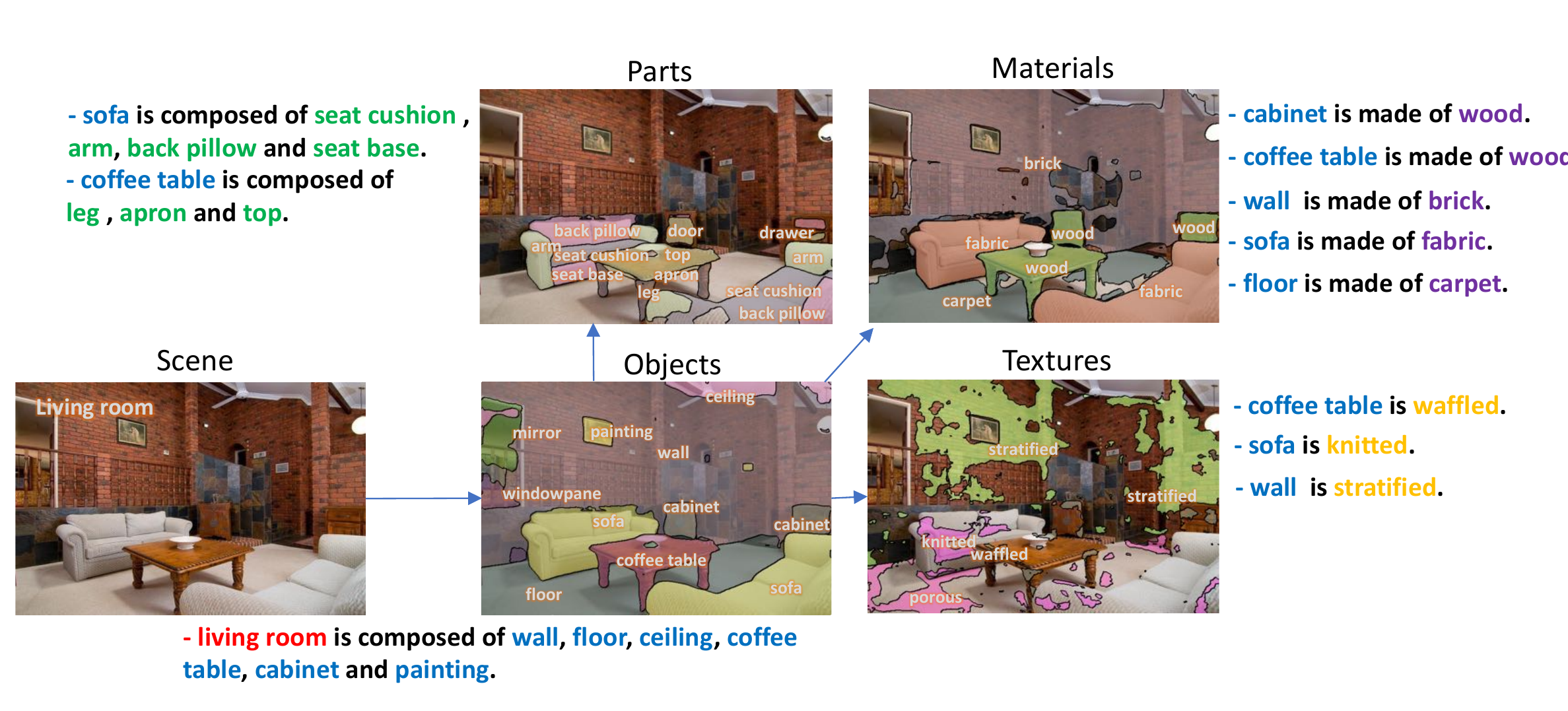}
\caption{Network trained for Unified Perceptual Parsing is able to parse various visual concepts at multiple perceptual levels such as scene, objects, parts, textures, and materials all at once. It also identifies the compositional structures among the detected concepts.}
\label{teaser}
\end{figure}

The human visual system is able to extract a remarkable amount of semantic information from a single glance. We not only instantly parse the objects contained within, but also identify the fine-grained attributes of objects, such as their parts, textures and materials. For example in Figure \ref{teaser}, we can recognize that this is a living room with various objects such as a coffee table, a painting, and walls inside. At the same time, we identify that the coffee table has legs, an apron and top, as well as that the coffee table is wooden and the surface of the sofa is knitted. Our interpretation of the visual scene is organized at multiple levels, from the visual perception of the materials and textures to the semantic perception of the objects and parts. 

Great progress in computer vision has been made towards human-level visual recognition because of the development of deep neural networks and large-scale image datasets. However, various visual recognition tasks are mostly studied independently. For example, human-level recognition has been reached for object classification \cite{he2016deep} and scene recognition \cite{zhou2017scene}; objects and stuff are parsed and segmented precisely at pixel-level \cite{hu2017learning,zhou2017scene}; Texture and material perception and recognition have been studied in \cite{cimpoi2014describing} and \cite{liu2010exploring}. Since scene recognition, object detection, texture and material recognition are intertwined in human visual perception, this raises an important question for the computer vision systems: is it possible for a neural network to solve several visual recognition tasks simultaneously? This motives our work to introduce a new task called Unified Perceptual Parsing (UPP) along with a novel learning method to address it. 

There are several challenges in UPP. First, there is no single image dataset annotated with all levels of visual information. Various image datasets are constructed only for specific task, such as ADE20K for scene parsing \cite{zhou2017scene}, the Describe Texture Dataset (DTD) for texture recognition \cite{cimpoi2014describing}, and OpenSurfaces for material and surface recognition \cite{bell2013opensurfaces}. Next, annotations from different perceptual levels are heterogeneous. For example, ADE20K has pixel-wise annotations while the annotations for textures in the DTD are image-level. 

To address the challenges above we propose a framework that overcomes the heterogeneity of different datasets and learns to detect various visual concepts jointly. On the one hand, at each iteration, we randomly sample a data source, and only update the related layers on the path to infer the concepts from the selected source. Such a design avoids erratic behavior that the gradient with respect to annotations of a certain concept may be noisy. On the other hand, our framework exploits the hierarchical nature of features from a single network, \ie, for concepts with higher-level semantics such as scene classification, the classifier is built on the feature map with the higher semantics only; for lower-level semantics such as object and material segmentation, classifiers are built on feature maps fused across all stages or the feature map with low-level semantics only. We further propose a training method that enables the network to predict pixel-wise texture labels using only image-level annotations. 

Our contributions are summarized as follows: 1) We present a new parsing task Unified Perceptual Parsing, which requires systems to parse multiple visual concepts at once. 2) We present a novel network called UPerNet with hierarchical structure to learn from heterogeneous data from multiple image datasets. 3) The model is shown to be able to jointly infer and discover the rich visual knowledge underneath images.

\subsection{Related work}

Our work is built upon the previous work of semantic segmentation and multi-task learning. 

\myparagraph{Semantic segmentation.} To generate pixel-wise semantic predictions for a given image, image classification networks \cite{krizhevsky2012imagenet,simonyan2014very,szegedy2015going,he2016deep} are extended to generate semantic segmentation masks. Pioneering work by Chen \etal~\cite{chen2014semantic}, based on structure prediction, uses conditional random field (CRF) to refine the activations of the final feature map of CNNs. The most prevalent framework designed for this pixel-level classification task is the Fully Convolutional Network (FCN)~\cite{long2015fully}, which replaces fully-connected layers in classification networks with convolutional layers. Noh \etal~\cite{noh2015learning} propose a framework which applies deconvolution~\cite{zeiler2011adaptive} to up-sample low resolution feature maps. Yu and Vladlen~\cite{yu2015multi} propose an architecture based on dilated convolution which is able to exponentially expand the receptive field without loss of resolution or coverage. More recently, RefineNet~\cite{lin2017refinenet} uses a coarse-to-fine architecture which exploits all information available along the down-sampling process. The Pyramid Scene Parsing Network (PSPNet)~\cite{zhao2017pyramid} performs spatial pooling at several grid scales and achieves remarkable performance on several segmentation benchmarks~\cite{everingham2010pascal,cordts2016cityscapes,zhou2017scene}. 

\myparagraph{Multi-task learning.} Multi-task learning, which aims to train models to accomplish multiple tasks at the same time, has attracted attention since long before the era of deep learning. For example, a number of previous research works focus on the combination of recognition and segmentation~\cite{keeler1991integrated,kokkinos2005expectation,maire2011object}. More recently, Elhoseiny \etal~\cite{elhoseiny2015convolutional} have proposed a model that performs pose estimation and object classification simultaneously. Eigen and Fergus~\cite{eigen2015predicting} propose an architecture that jointly addresses depth prediction, surface normal estimation, and semantic labeling. Teichmann \etal~\cite{teichmann2016multinet} propose an approach to perform classification, detection, and semantic segmentation via a shared feature extractor. Kokkinos proposes the UberNet~\cite{kokkinos2016ubernet}, a deep architecture that is able to do seven different tasks relying on diverse training sets. Another recent work \cite{hu2017learning} proposes a partially supervised training paradigm to scale up the segmentation of objects to $3,000$ objects using box annotations only. Comparing our work with previous works on multi-task learning, only a few of them perform multi-task learning on heterogeneous datasets, \ie, a dataset that does not necessarily have all levels of annotations over all tasks. Moreover, although tasks in~\cite{kokkinos2016ubernet} are formed from low level to high level, such as boundary detection, semantic segmentation and object detection, these tasks do not form the hierarchy of visual concepts. In Section~\ref{sec:visual_knowledge}, we further demonstrate the effectiveness of our proposed tasks and frameworks in discovering the rich visual knowledge from images.

\section{Defining Unified Perceptual Parsing}

We define the task of Unified Perceptual Parsing as the recognition of many visual concepts as possible from a given image. Possible visual concepts are organized into several levels: from scene labels, objects, and parts of objects, to materials and textures of objects. The task depends on the availability of different kinds of training data. Since there is no single image dataset annotated with all visual concepts at multiple levels, we first construct an image dataset by combining several sources of image annotations. 

\subsection{Datasets}

In order to accomplish segmentation of a wide range of visual concepts from multiple levels, we utilize the Broadly and Densely Labeled Dataset (Broden)~\cite{netdissect2017}, a heterogeneous dataset that contains various visual concepts. Broden unifies several densely labeled image datasets, namely ADE20K~\cite{zhou2017scene}, Pascal-Context~\cite{mottaghi_cvpr14}, Pascal-Part~\cite{chen_cvpr14}, OpenSurfaces~\cite{bell2013opensurfaces}, and the Describable Textures Dataset (DTD)~\cite{cimpoi2014describing}. These datasets contain samples of a broad range of scenes, objects, object parts, materials and textures in a variety of contexts. Objects, object parts and materials are segmented down to pixel level while textures and scenes are annotated at image level. 

The Broden dataset provides a wide range of visual concepts. Nevertheless, since it is originally collected to discover the alignment between visual concepts and hidden units of Convolutional Neural Networks (CNNs) for network interpretability~\cite{netdissect2017,netdissect2018}, we find that samples from different classes are unbalanced. Therefore we standardize the Broden dataset to make it more suitable for training segmentation networks. First, we merge similar concepts across different datasets. For example, objects and parts annotations in ADE20K, Pascal-Context, and Pascal-Part are merged and unified. Second, we only include object classes which appear in at least 50 images \emph{and} contain at least $50,000$ pixels in the whole dataset. Also, object parts which appear in at least 20 images can be considered valid parts. Objects and parts that are conceptually inconsistent are manually removed. Third, we manually merge under-sampled labels in OpenSurfaces. For example, \textit{stone} and \textit{concrete} are merged into \textit{stone}, while \textit{clear plastic} and \textit{opaque plastic} are merged into \textit{plastic}. Labels that appear in less than $50$ images are also filtered out. Fourth, we map more than 400 scene labels from the ADE20K dataset to $365$ labels from the Places dataset~\cite{zhou2014learning}. 

\begin{figure}[!t]
\centering
\includegraphics[width=1.0\linewidth]{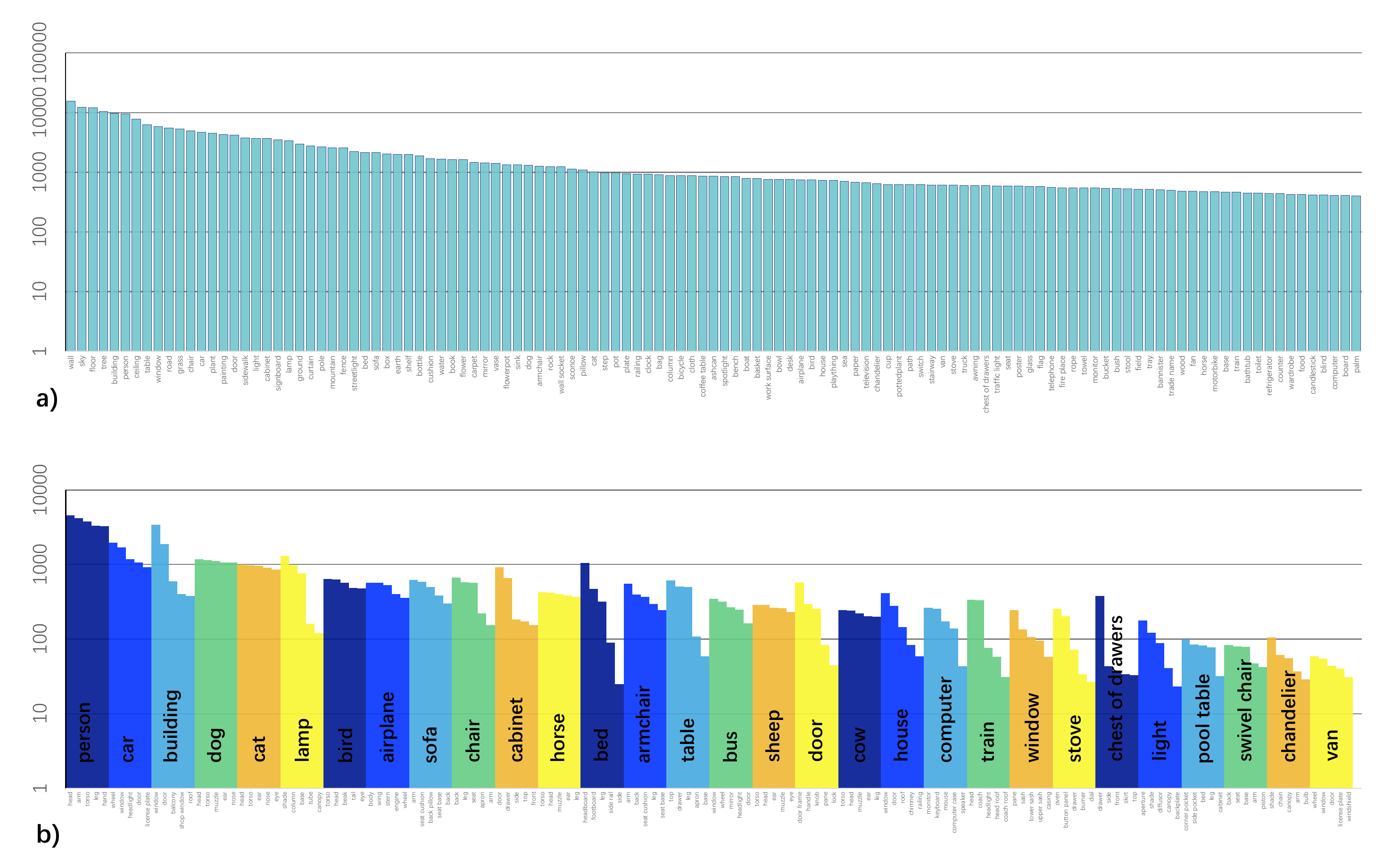}
\caption{a) Sorted object classes by frequency: we show top 120 classes selected from the Broden+. Object classes that appear in less than 50 images or contain less than $50,000$ pixels are filtered. b) Frequency of parts grouped by objects. We show only top 30 objects with their top 5 frequent parts. The parts that appear in less than $20$ images are filtered.}
\label{fig:object_part_statistics}
\end{figure}

\begin{figure}[!t]
\centering
\includegraphics[width=1.0\linewidth]{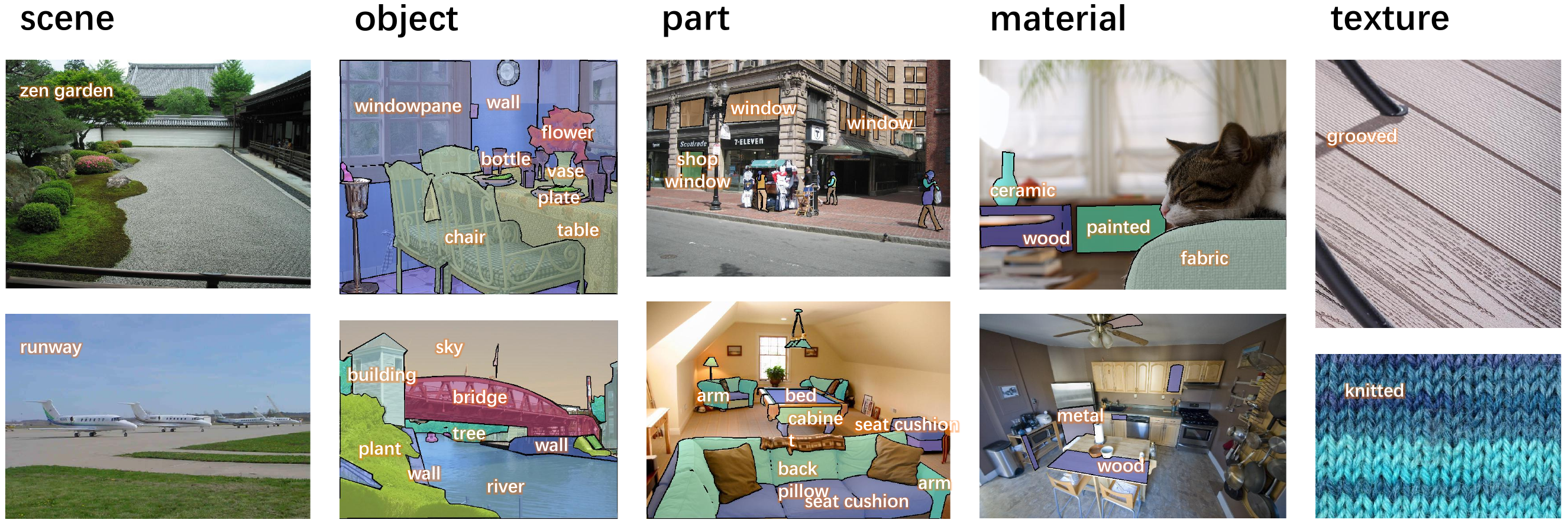}
\caption{Samples from the Broden+ dataset. The ground-truth labels for scene and texture are image-level annotations, while for object, part and material are pixel-wise annotations. Object and part are densely annotated, while material is partially annotated. Images with texture labels are mostly such localized object regions.}
\label{fig:ground_truth_examples}
\end{figure}

Table~\ref{tab:modified_broden_label_statistics} shows some statistics of our standardized Broden, termed as Broden+. It contains $57,095$ images in total, including $22,210$ images from ADE20K, $10,103$ images from Pascal-Context and Pascal-Part, $19,142$ images from OpenSurfaces and $5,640$ images from DTD. Figure~\ref{fig:object_part_statistics} shows the distribution of objects as well as parts grouped by the objects to which they belong. We also provide examples from each source of the Broden+ dataset in Figure~\ref{fig:ground_truth_examples}. 

\begin{table}[!tbp]
\begin{center}
\setlength{\tabcolsep}{5.8pt}
\begin{tabular}{ l r l l}
\toprule[0.8pt]
\textbf{Category} & \textbf{Classes} & \textbf{Sources} & \textbf{Eval. Metrics} \\
\hline
scene & 365 & ADE~\cite{zhou2017scene} & top-1 acc. \\
object & 335 & ADE~\cite{zhou2017scene}, Pascal-Context\cite{mottaghi_cvpr14} & mIoU \& pixel acc. \\
object w/ part & 77 & ADE~\cite{zhou2017scene}, Pascal-Context\cite{mottaghi_cvpr14} & - \\
part & 152 & ADE~\cite{zhou2017scene}, Pascal-Part~\cite{chen_cvpr14} & mIoU (bg) \& pixel acc. \\
material & 26 & OpenSurfaces~\cite{bell2013opensurfaces} & mIoU \& pixel acc. \\
texture & 47 & DTD~\cite{cimpoi2014describing} & top-1 acc. \\
\bottomrule[0.8pt]
\end{tabular}
\end{center}
\caption{Statistics of each label type in the Broden+ dataset. Evaluation metrics for each type of labels are also listed.}
\label{tab:modified_broden_label_statistics}
\end{table}

\subsection{Metrics}
\label{sec:metrics}

To quantify the performance of models, we set different metrics based on the annotations of each dataset. Standard metrics to evaluate semantic segmentation tasks include Pixel Accuracy (P.A.), which indicates the proportion of correctly classified pixels, and mean IoU (mIoU), which indicates the intersection-over-union (IoU) between the predicted and ground truth pixels, averaged over all object classes. Note that since there might be unlabeled areas in an image, the mIoU metric will not count the predictions on unlabeled regions. This would encourage people to exclude the background label during training. However, it is not suitable for the evaluation of tasks like part segmentation, because for some objects the regions with part annotations only account for a small number of pixels. Therefore we use mIoU, but count the predictions in the background regions, denoted as mIoU-bg, in certain tasks. In this way, excluding background labels during training will boost P.A. by a small margin. Nonetheless, it will significantly downgrade mIoU-bg performance. 

For object and material parsing involving ADE20K, Pascal-Context, and OpenSurfaces, the annotations are at pixel level. Images in ADE20K and Pascal-Context are fully annotated, with the regions that do not belong to any pre-defined classes categorized into an unlabeled class. Images in OpenSurfaces are partially annotated, \ie, if several regions of material occur in a single image, more than one region may not be annotated. We use P.A. and mIoU metrics for these two tasks. 

For object parts we use P.A. and mIoU-bg metrics for the above mentioned reason. The IoU of each part is first averaged within an object category, then averaged over all object classes. For scene and texture classification we report top-1 accuracy. Evaluation metrics are listed in Table~\ref{tab:modified_broden_label_statistics}.

To balance samples across different labels in different categories we first randomly sample $10\%$ of original images as the validation set. We then randomly choose an image both from the training and validation set, and check if the annotations in pixel level are more balanced towards $10\%$ after swapping these two images. The process is performed iteratively. The dataset is split into $51,617$ images for training and $5,478$ images for validation.

\section{Designing Networks for Unified Perceptual Parsing}
\label{sec:framework}

\begin{figure}[!t]
\centering
\includegraphics[width=1.0\linewidth]{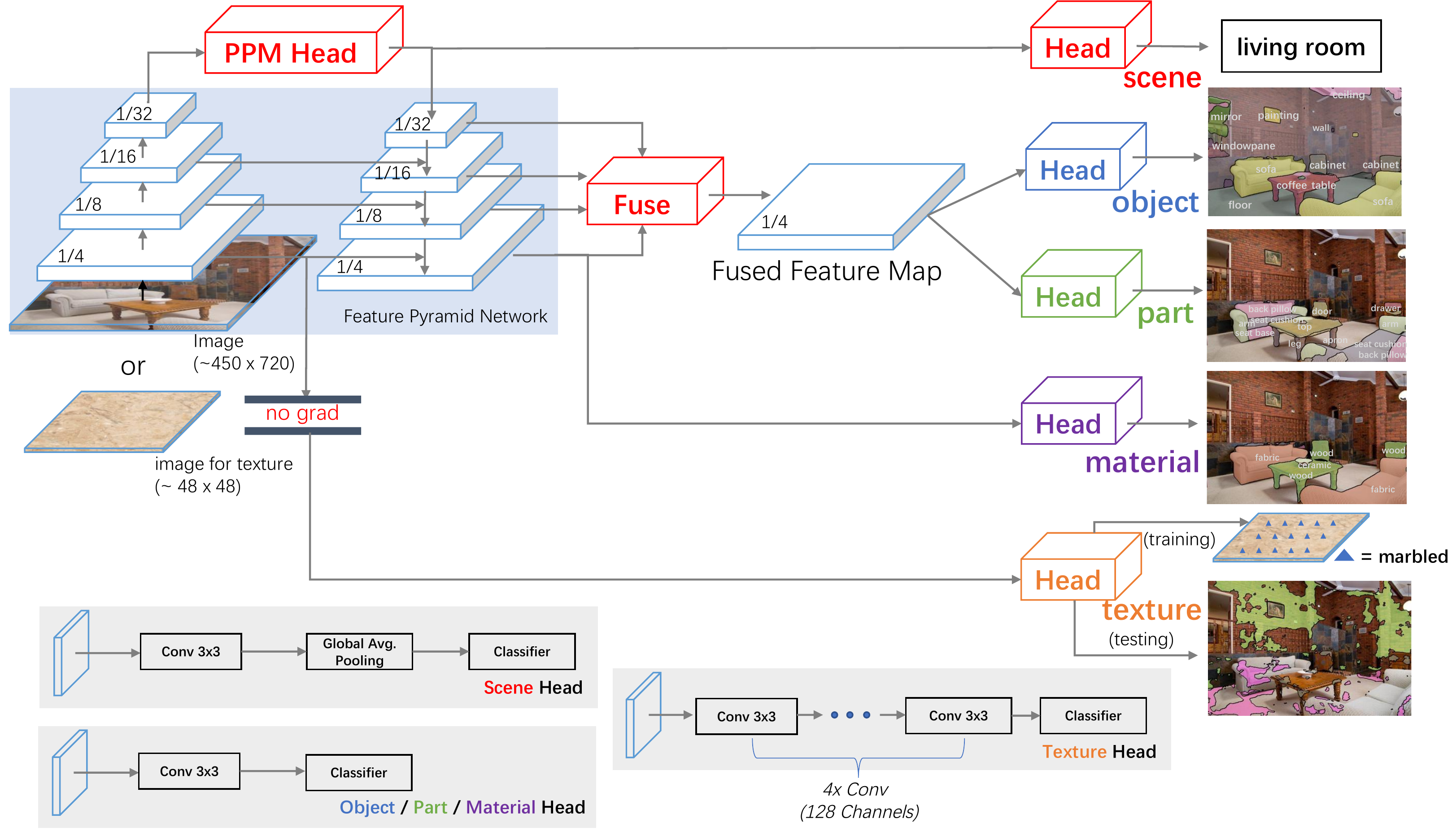}
\caption{UPerNet framework for Unified Perceptual Parsing. Top-left: The Feature Pyramid Network (FPN)~\cite{lin2017feature} with a Pyramid Pooling Module (PPM)~\cite{zhao2017pyramid} appended on the last layer of the back-bone network before feeding it into the top-down branch in FPN. Top-right: We use features at various semantic levels. Scene head is attached on the feature map directly after the PPM since image-level information is more suitable for scene classification. Object and part heads are attached on the feature map fused by all the layers put out by FPN. Material head is attached on the feature map in FPN with the highest resolution. Texture head is attached on the Res-2 block in ResNet~\cite{he2016deep}, and fine-tuned after the whole network finishes training on other tasks. Bottom: The illustrations of different heads. Details can be found in Section~\ref{sec:framework}.}
\label{fig:framework}
\end{figure}

We demonstrate our network design in Figure~\ref{fig:framework}, termed as \textbf{UPerNet} (\textbf{U}nified \textbf{Per}ceptual Parsing \textbf{Net}work), based on the Feature Pyramid Network (FPN)~\cite{lin2017feature}. FPN is a generic feature extractor which exploits multi-level feature representations in an inherent and pyramidal hierarchy. It uses a top-down architecture with lateral connections to fuse high-level semantic information into middle and low levels with marginal extra cost. To overcome the issue raised by Zhou~\etal~ \cite{zhou2014object} that although the theoretical receptive field of deep CNN is large enough, the empirical receptive field of deep CNN is relatively much smaller~\cite{zhou2016learning}, we apply a Pyramid Pooling Module (PPM) from PSPNet~\cite{zhao2017pyramid} on the last layer of the backbone network before feeding it into the top-down branch in FPN. Empirically we find that the PPM is highly compatible with the FPN architecture by bringing effective global prior representations. For further details on FPN and PPM, we refer the reader to~\cite{lin2017feature} and~\cite{zhao2017pyramid}.

With the new framework, we are able to train a single network which is able to unify parsing of visual attributes at multiple levels. Our framework is based on Residual Networks~\cite{he2016deep}. We denote the set of last feature maps of each stage in ResNet as $\left\{ C_2, C_3, C_4, C_5\right\}$, and the set of feature maps put out by FPN as $\left\{ P_2, P_3, P_4, P_5 \right\}$, where $P_5$ is also the feature map directly following PPM. The down-sampling rates are $\left\{ 4, 8, 16, 32\right\}$, respectively. \emph{Scene label}, the highest-level attribute annotated at image-level, is predicted by a global average pooling of $P_5$ followed by a linear classifier. It is worth noting that, unlike frameworks based on a dilated net, the down-sampling rate of $P_5$ is relatively large so that the features after global average pooling focus more on high-level semantics. For \emph{object label}, we empirically find that fusing all feature maps of FPN is better than only using the feature map with the highest resolution ($P_2$). \emph{Object parts} are segmented based on the same feature map as objects. For \emph{materials}, intuitively, if we have prior knowledge that these areas belong to the object ``cup", we are able to make a reasonable conjecture that it might be made up of paper or plastics. This context is useful, but we still need local apparent features to decide which one is correct. It should also be noted that an object can be made up of various materials. Based on the above observations, we segment materials on top of $P_2$ rather than fused features. \emph{Texture label}, given at the image-level, is based on non-natural images. Directly fusing these images with other natural images is harmful to other tasks. Also we hope the network can predict texture labels at pixel level. To achieve such a goal, we append several convolutional layers on top of $C_2$, and force the network to predict the texture label at every pixel. The gradient of this branch is prevented from back-propagating to layers of backbone networks, and the training images for texture are resized to a smaller size ($\sim{}64\times{}64$). The reasons behind these designs are: 1) Texture is the lowest-level perceptual attribute, thus it is purely based on apparent features and does not need any high-level information. 2) Essential features for predicting texture correctly are implicitly learned when trained on other tasks. 3) The receptive field of this branch needs to be small enough, so that the network is able to predict different labels at various regions when an image at normal scale is fed in the network. We only fine-tune the texture branch for a few epochs after the whole network finishes training on other tasks.

When only trained on object supervision, without further enhancements, our framework yields almost identical performance as the state-of-the-art PSPNet, while requiring only $63\%$ of training time for the same number of epochs. It is worth noting that we do not even perform deep supervision or data augmentations used in PSPNet other than scale jitter, according to the experiments in their paper~\cite{zhao2017pyramid}. Ablation experiments are provided in Section~\ref{sec:results}.

\subsection{Implementation details}
Every classifier is preceded by a separate convolutional head. To fuse the layers with different scales such as $\left\{ P_2, P_3, P_4, P_5 \right\}$, we resize them via bilinear interpolation to the size of $P_2$ and concatenate these layers. A convolutional layer is then applied to fuse features from different levels as well as to reduce channel dimensions. All extra non-classifier convolutional layers, including those in FPN, have batch normalization~\cite{ioffe2015batch} with $512$-channel output. ReLU~\cite{nair2010rectified} is applied after batch normalization. Same as~\cite{chen2016deeplab}, we use the ``poly'' learning rate policy where the learning rate at current iteration equals the initial learning rate multiplying $\left( 1 - \frac{iter}{max\_iter}\right)^{power}$. The initial learning rate and power are set to $0.02$ and $0.9$, respectively. We use a weight decay of $0.0001$ and a momentum of $0.9$. During training the input image is resized such that the length of its shorter side is randomly chosen from the set $\left\{300, 375, 450, 525, 600\right\}$. For inference we do not apply multi-scale testing for fair comparison, and the length is set to $450$. The maximum length of the longer side is set to $1200$ in avoidance of GPU memory overflow. The layers in the backbone network are initialized with weights pre-trained on ImageNet~\cite{deng2009imagenet}. 

During each iteration, if a mini-batch is composed of images from several sources on various tasks, the gradient with respect to a certain task can be noisy, since the real batch size of each task is in fact decreased. Thus we randomly sample a data source at each iteration based on the scale of each source, and only update the path to infer the concepts related to the selected source. For object and material, we do not calculate loss on unlabeled area. For part, as mentioned in Section~\ref{sec:metrics}, we add background as a valid label. Also the loss of a part is applied only inside the regions of its super object.

Due to physical memory limitations a mini-batch on each GPU involves only $2$ images. We adopt synchronized SGD training across $8$ GPUs. It is worth noting that batch size has proven to be important to generate accurate statistics for tasks like classification~\cite{ioffe2017batch}, semantic segmentation~\cite{zhao2017pyramid} and object detection~\cite{peng2017megdet}. We implement batch normalization such that it is able to synchronize across multiple GPUs. We do not fix any batch norm layer during training. The number of training iterations of ADE20k (with $\sim{}20k$ images) alone is $100k$. If trained on a larger dataset, we linearly increase training iterations based on the number of images in the dataset.

\subsection{Design discussion}
State-of-the-art segmentation networks are mainly based on fully convolutional networks (FCNs)~\cite{long2015fully}. Due to a lack of sufficient training samples, segmentation networks are usually initialized from networks pre-trained for image classification~\cite{deng2009imagenet,krizhevsky2012imagenet,simonyan2014very}. To enable high-resolution predictions for semantic segmentation, dilated convolution~\cite{yu2015multi}, a technique which removes the stride of convolutional layers and adds holes between each location of convolution filters, has been proposed to ease the side effect of down-sampling while maintaining the expansion rate for receptive fields. The dilated network has become the \textit{de facto} paradigm for semantic segmentation.

We argue that such a framework has major drawbacks for the proposed Unified Perceptual Parsing task. First, recently proposed deep CNNs~\cite{he2016deep,xie2017aggregated}, which have succeeded on tasks such as image classification and semantic segmentation usually have tens or hundreds of layers. These deep CNNs are intricately designed such that the down-sampling rate grows rapidly in the early stage of the network for the sake of a larger receptive field and lighter computational complexity. For example, in the ResNet with $100$ convolutional layers in total, there are $78$ convolutional layers in the Res-4 and Res-5 blocks combined, with down-sampling rates of $16$ and $32$, respectively. In practice, in a dilated segmentation framework, dilated convolution needs to be applied to both blocks to ensure that the maximum down-sampling rate of all feature maps do not exceed 8. Nevertheless, due to the feature maps within the two blocks are increased to $4$ or $16$ times of their designated sizes, both the computation complexity and GPU memory footprint are dramatically increased. The second drawback is that such a framework utilizes only the deepest feature map in the network. Prior works~\cite{zeiler2014visualizing} have shown the hierarchical nature of the features in the network, \ie, lower layers tend to capture local features such as corners or edge/color conjunctions, while higher layers tend to capture more complex patterns such as parts of some object. Using the features with the highest-level semantics might be reasonable for segmenting high-level concepts such as objects, but it is naturally unfit to segment perceptual attributes at multiple levels, especially the low-level ones such as textures and materials. In what follows, we demonstrate the effectiveness and efficiency of our UPerNet.

\begin{table}[!tbp]
\begin{center}
\setlength{\tabcolsep}{5.8pt}
\begin{tabular}{l r r r | r}
\toprule[0.8pt]
Method & Mean IoU(\%) & Pixel Acc.(\%) & Overall(\%) & Time(hr) \\
\hline
FCN~\cite{long2015fully} & 29.39 & 71.32 & 50.36 & - \\
SegNet~\cite{badrinarayanan2017segnet} & 21.64 & 71.00 & 46.32 & - \\
DilatedNet~\cite{yu2015multi} & 32.31 & 73.55 & 52.93 & - \\
CascadeNet~\cite{zhou2017scene} & 34.90 & 74.52 & 54.71 & - \\
RefineNet (Res-152)~\cite{lin2017refinenet} & 40.70 & - & - & - \\
DilatedNet${}^{*}{}^{\dagger}$(Res-50)~\cite{zhao2017pyramid} & 34.28 & 76.35 & 55.32 & 53.9 \\
PSPNet${}^{\dagger}$(Res-50)~\cite{zhao2017pyramid} & \textcolor{blue}{\textbf{41.68}} & \textcolor{blue}{\textbf{80.04}} & \textcolor{blue}{\textbf{60.86}} & 61.1 \\
\hline
FPN ($/16$) & 34.46 & 76.04 & 55.25 & 18.1 \\
FPN ($/8$) & 34.99 & 76.54 & 55.77 & 20.2 \\
FPN ($/4$) & 35.26 & 76.52 & 55.89 & 21.2 \\
FPN+PPM ($/4$) & 40.13 & 79.61 & 59.87 & 27.8 \\
FPN+PPM+Fusion ($/4$) & \textcolor{red}{\textbf{41.22}} & \textcolor{red}{\textbf{79.98}} & \textcolor{red}{\textbf{60.60}} & 38.7 \\
\bottomrule[0.8pt]
\end{tabular}
\end{center}
\caption{Detailed analysis of our framework based on ResNet-50 \emph{v.s.} state-of-the-art methods on ADE20K dataset. Our results are obtained without multi-scale inference or other techniques. FPN baseline is competitive while requiring much less computational resources. Further increasing resolution of feature maps brings consistent gain. PPM is highly compatible with FPN. Empirically we find that fusing features from all levels of FPN yields best performance. ${}^{*}$: A stronger reference for DilatedNet reported in~\cite{zhao2017pyramid}. ${}^{\dagger}$: Training time is based on our reproduced models. We also use the same codes in FPN baseline.}
\label{tab:framework_on_ade}
\end{table}

\section{Experiments}
The experiment section is organized as follows: we first introduce the quantitative study of our proposed framework on the original semantic segmentation task and the UPP task in Section~\ref{sec:results}. Then we apply the framework to discover visual common sense knowledge underlying scene understanding in Section~\ref{sec:visual_knowledge}. 

\subsection{Main results}
\label{sec:results}

\myparagraph{Overall architecture.} To demonstrate the effectiveness of our proposed architecture on semantic segmentation, we report the results trained on ADE20K using object annotations under various settings in Table~\ref{tab:framework_on_ade}. In general, FPN demonstrates competitive performance while requiring much less computational resources for semantic segmentation. Using the feature map up-sampled only once with a down-sampling rate of $16$ ($P_4$), it reaches mIoU and P.A. of $34.46/76.04$, almost identical to the strong baseline reference reported in~\cite{zhao2017pyramid} while only taking about $1/3$ of the training time for the same number of iterations. Performance improves further when the resolution is higher. Adding the Pyramid Pooling Module (PPM) boosts performance by a $4.87/3.09$ margin, which demonstrates that FPN also suffers from an insufficient receptive field. Empirically we find that fusing features from all levels of FPN yields best performance, a consistent conclusion also observed in~\cite{fair2017cocoseg}.

\begin{figure}[!t]
\centering
\includegraphics[width=0.9\linewidth]{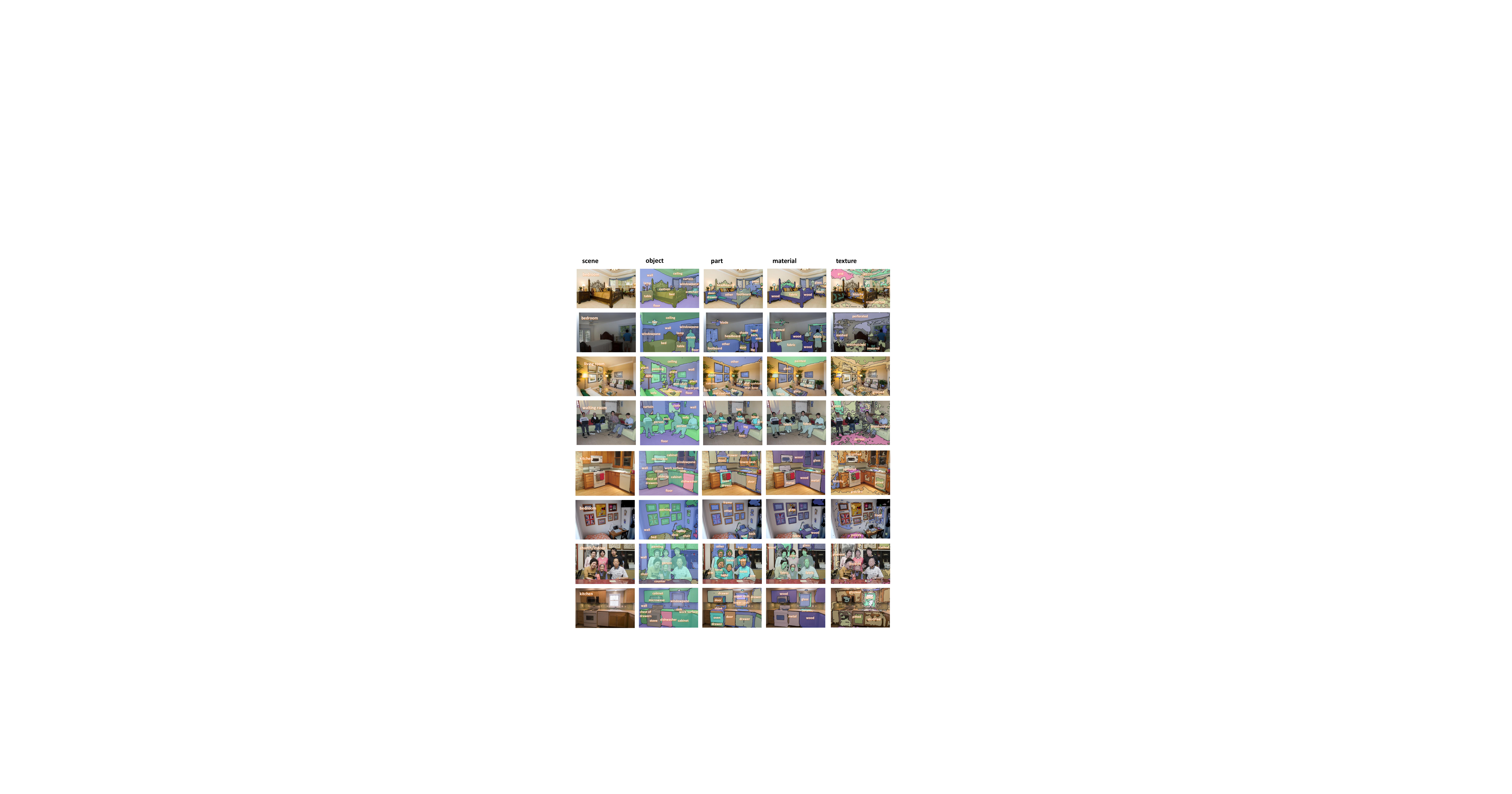}
\caption{Predictions on the validation set using UPerNet (ResNet-50). From left to right: scene classification, and object, part, material, and texture parsing.}
\label{fig:visualization_samples}
\end{figure}

The performance of FPN is surprising considering its simplicity with feature maps being simply up-sampled by bilinear interpolation instead of time-consuming deconvolution, and the top-down path is fused with bottom-up path by an 1x1 convolutional layer followed by element-wise summation without any complex refinement module. It is the simplicity that accomplishes its efficiency. We therefore adopt this design for Unified Perceptual Parsing.

\begin{table*}[!t]
\begin{center}
\setlength{\tabcolsep}{1mm}{
\begin{tabular}{ccccc|rr rr r rr r}
\toprule[0.8pt]
\multicolumn{5}{l|}{Training Data} & \multicolumn{2}{c}{Object} & \multicolumn{2}{c}{Part} & Scene & \multicolumn{2}{c}{Material} & Texture \\
\hline
 		+O & +P & +S & +M & +T & ~mI.~ & ~P.A.~ & mI.(bg) & ~P.A.~ & ~T-1~ & ~mI.~ & ~P.A.~ & ~T-1~ \\

 		\hline
 		$\checkmark$ &  &  &  &  & 24.72 & 78.03 & - & - & - & - & - & - \\
         &  &  & $\checkmark$ &  & - & - & - & - & - & 52.78 & 84.32 & - \\
         \hline
         $\checkmark$ & $\checkmark$ &  &  &  & 23.92 & 77.48 & 30.21 & 48.30 & - & - & - & - \\
         $\checkmark$ & $\checkmark$ & $\checkmark$ &  &  & 23.83 & 77.23 & 30.10 & 48.34 & 71.35 & - & - & - \\
         $\checkmark$ & $\checkmark$ & $\checkmark$ & $\checkmark$ &  & 23.36 & 77.09 & 28.75 & 46.92 & 70.87 & 54.19 & 84.45 & - \\
         \hline
         $\checkmark$ & $\checkmark$ & $\checkmark$ & $\checkmark$ & $\checkmark$ & 23.36 & 77.09 & 28.75 & 46.92 & 70.87 & 54.19 & 84.45 & 35.10 \\
         \bottomrule[0.8pt]
 		\end{tabular}}
 	\end{center}
 	\caption{Results of Unified Perceptual Parsing on the Broden+ dataset. O: Object. P: Part. S: Scene. M: Material. T: Texture. mI.: mean IoU. P.A.: pixel accuracy. mI.(bg): mean IoU including background. T-1: top-1 accuracy. }
 	\label{tab:main_results}
\end{table*}

\begin{figure}[!t]
\centering
\subfigure[Visualization of scene-object relations. Indoor scenes and outdoor scenes are clustered into different groups (left part of top image and right part of top image). We are also able to locate a common object appearing in various scenes, or find the objects in a certain scene (bottom left and bottom right).]{\includegraphics[width=0.9\linewidth]{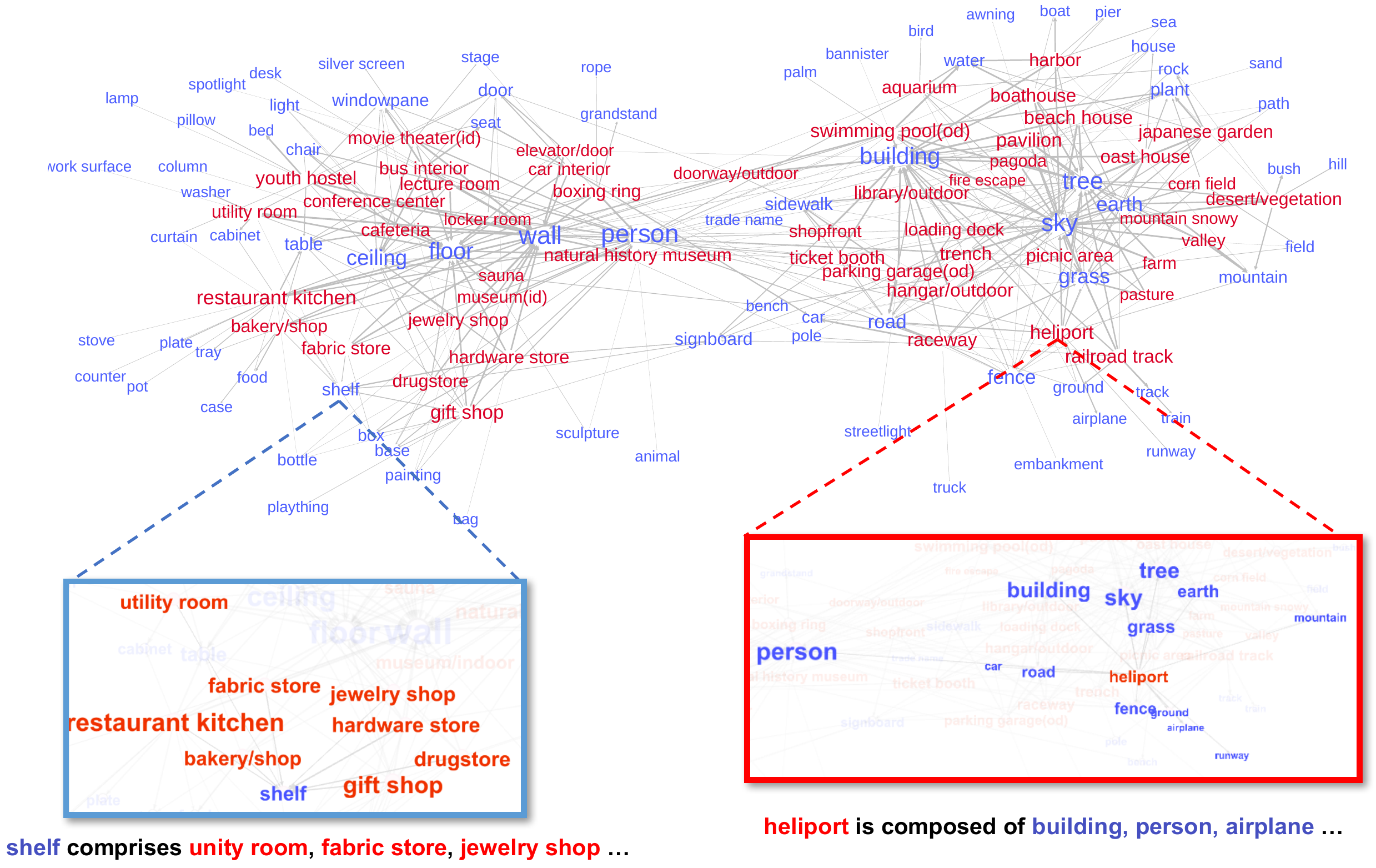} \label{fig:scene_object_relation}}
\subfigure[From left to right: visualizations of object-material relations, part-material relations and material-texture relations. We are able to discover knowledge such as some sinks are ceramic while others are metallic. We can also find out what can be used to describe a material.]{\includegraphics[width=1.0\linewidth]{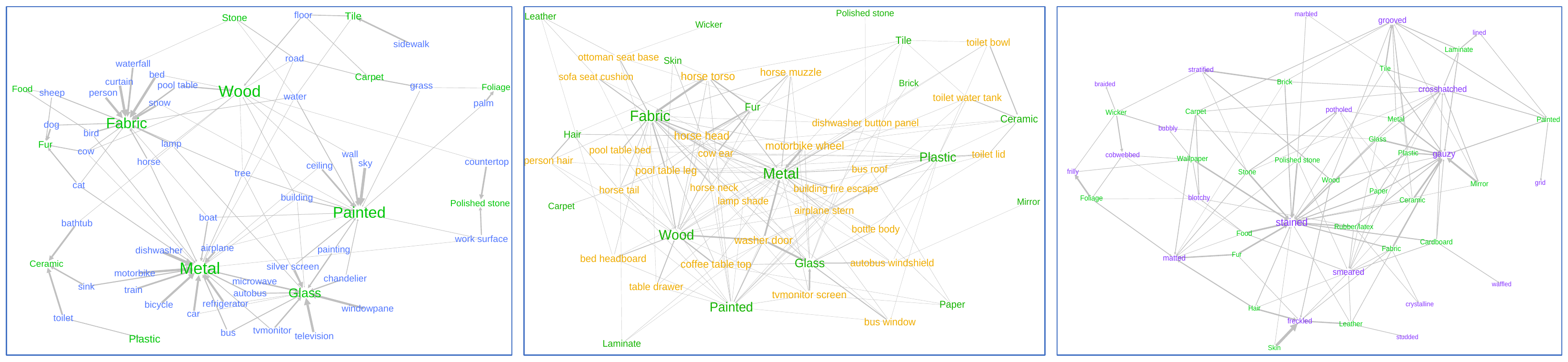} \label{fig:other_relations}}
\caption{Visualizing discovered compositional relations between various concepts.}
\end{figure}

\myparagraph{Multi-task learning with heterogeneous annotations.} We report the results trained on separate or fused different sets of annotations. The baseline of object parsing is the model trained on ADE20K and Pascal-Context. It yields mIoU and P.A. of $24.72/78.03$. This result, compared with the results for ADE20K, is relatively low because Broden+ has many more object classes. The baseline of material is the model trained on OpenSurfaces. It yields mIoU and P.A. of $52.78/84.32$. Joint training of object and part parsing yields $23.92/77.48$ on object and $30.21/48.30$ on part. The performance on object parsing trained plus part annotations is almost identical to that trained only on object annotations. After adding a scene prediction branch it yields top-1 accuracy of $71.35\%$ on scene classification, with negligible downgrades of object and part performance. When jointly training material with object, part, and scene classification, it yields a performance of $54.19/84.45$ on material parsing, $23.36/77.09$ on object parsing, and $28.75/46.92$ on part parsing. It is worth noting that the object and part both suffer a slight performance degrade due to heterogeneity, while material enjoys a boost in performance compared with that trained only on OpenSurfaces. We conjecture that it is attributed to the usefulness of information in object as priors for material parsing. As mentioned above, we find that directly fusing texture images with other natural images is harmful to other tasks, since there are nontrivial differences between images in DTD and natural images. After fine-tuning on texture images using the model trained with all other tasks, we can obtain the quantitative texture classification results by picking the most frequent pixel-level predictions as an image-level prediction. It yields classification accuracy of $35.10$. The performance on texture indicates that only fine-tuning the network on texture labels is not optimal. However, this is a necessary step to overcome the fusion of natural and synthetic data sources. We hope future research can discover ways to better utilize such image-level annotations for pixel-level predictions.

\myparagraph{Qualitative results.} We provide qualitative results of UPerNet, as visualized in Figure~\ref{fig:visualization_samples}. UPerNet is able to unify compositional visual knowledge and efficiently predicts hierarchical outputs simultaneously.

\subsection{Discovering visual knowledge in natural scenes}
\label{sec:visual_knowledge}

Unified Perceptual Parsing requires a model that is able to recognize as many visual concepts as possible from a given image. If a model successfully achieves this goal, it could discover rich visual knowledge underlying the real world, such as answering questions like ``What are the commonalities between living rooms and bedrooms?'' or ``What are the materials that make a cup?'' The discovery or even the reasoning of visual knowledge in natural scenes will enable future vision systems to understand its surroundings better. In this section, we demonstrate that our framework trained on the Broden+ is able to discover compositional visual knowledge at multiple levels. That is also the special application for the network trained on heterogeneous data annotations. We use the validation set of Places-365~\cite{zhou2014learning} containing $36,500$ images from 365 scenes as our testbed, since the Places dataset contains images from a variety of scenes and is closer to real world. We define several relations in a hierarchical way, namely \emph{scene-object} relation, \emph{object-part} relation, \emph{object-material} relation, \emph{part-material} relation and \emph{material-texture} relation. Note that only the object-part relations can be directly read out from the ground-truth annotations, other types of relations can only be extracted from the network predictions. 

\begin{table*}[!ht]
\begin{center}
\setlength{\tabcolsep}{5.8pt}{
\begin{tabular}{l}
\toprule[0.8pt]
\multicolumn{1}{c}{\textbf{Scene-object Relations}} \\
\textcolor{red}{garage (indoor)} is composed of \textcolor{blue}{floor}, \textcolor{blue}{wall}, \textcolor{blue}{ceiling}, \textcolor{blue}{car}, \textcolor{blue}{door}, \textcolor{blue}{person}, \textcolor{blue}{building},\\ \textcolor{blue}{windowpane}, \textcolor{blue}{box}, and \textcolor{blue}{signboard}.\\

\textcolor{red}{glacier} is composed of \textcolor{blue}{mountain}, \textcolor{blue}{sky}, \textcolor{blue}{earth}, \textcolor{blue}{tree}, \textcolor{blue}{snow}, \textcolor{blue}{rock}, \textcolor{blue}{water}, and \textcolor{blue}{person}.\\

\textcolor{red}{laundromat} is composed of \textcolor{blue}{wall}, \textcolor{blue}{floor}, \textcolor{blue}{washer}, \textcolor{blue}{ceiling}, \textcolor{blue}{door}, \textcolor{blue}{cabinet}, \textcolor{blue}{person}, \\ \textcolor{blue}{table} and \textcolor{blue}{signboard}.\\
\hline
\multicolumn{1}{c}{\textbf{Object-material Relations}} \\
\textcolor{blue}{toilet} is made of \textcolor{purple}{ceramic} (65\%) and \textcolor{purple}{plastic} (35\%). \\

\textcolor{blue}{microwave} is made of \textcolor{purple}{glass} (55\%), and \textcolor{purple}{metal} (45\%). \\

\textcolor{blue}{sidewalk} is made of \textcolor{purple}{tile} (65\%), \textcolor{purple}{stone} (18\%), and \textcolor{purple}{wood} (17\%). \\

\hline

\multicolumn{1}{c}{\textbf{Part-material Relations}} \\
\textcolor{green}{coffee table top} is made of \textcolor{purple}{wood} (69\%) and \textcolor{purple}{glass} (31\%). \\

\textcolor{green}{bed headboard} is made of \textcolor{purple}{wood} (77\%) and \textcolor{purple}{fabric} (23\%). \\

\textcolor{green}{tv monitor screen} is made of \textcolor{purple}{glass} (100\%).\\

\hline

\multicolumn{1}{c}{\textbf{Material-texture Relations}} \\
\textcolor{purple}{brick} is \textcolor{orange}{stratified} (42\%), \textcolor{orange}{stained} (34\%) and \textcolor{orange}{crosshatched} (24\%) . \\

\textcolor{purple}{stone} is \textcolor{orange}{stained} (43\%), \textcolor{orange}{potholed} (31\%) and \textcolor{orange}{matted} (26\%) . \\

\textcolor{purple}{mirror} is \textcolor{orange}{gauzy} (54\%), \textcolor{orange}{crosshatched} (26\%) and \textcolor{orange}{grooved} (20\%) . \\
\bottomrule[0.8pt]
\end{tabular}}
\end{center}
\caption{Discorved visual knowledge by UPerNet trained for UPP. UPerNet is able to extract reasonable visual knowledge priors.}
\label{tab:auto_generation}
\end{table*}


\myparagraph{Scene-object relations.} For each scene, we count how many objects show up normalized by the frequency of this scene. According to~\cite{brandes2013network}, we formulate the relation as a bipartite graph $G=(V, E)$ comprised of a set $V=V_s \cup V_o$ of scene nodes and object nodes together with a set $E$ of edges. The edge with a weight from $v_s$ to $v_o$ represents the percent likelihood that object $v_o$ shows up in scene $v_s$. No edge connects two nodes that are both from $V_s$ or both from $V_o$. We filter the edges whose weight is lower than a threshold and run a clustering algorithm to form a better layout. Due to space limitations, we only sample dozens of nodes and show the visualization of the graph in Figure~\ref{fig:scene_object_relation}. We can clearly see hat the indoor scenes mostly share objects such as ceiling, floor, chair, or windowpane while the outdoor scenes mostly share objects such as sky, tree, building, or mountain. What is more interesting is that even in the set of scenes, human-made and natural scenes are clustered into different groups. In the layout, we are also able to locate a common object appearing in various scenes, or find the objects in a certain scene. The bottom-left and bottom-right pictures in Figure~\ref{fig:scene_object_relation} illustrate an example in which we can reasonably conclude that the shelf often appears in shops, stores, and utility rooms; and that in a heliport there are often trees, fences, runways, persons, and of course, airplanes.

\myparagraph{Object(part)-material relations.} Apart from scene-object relations, we are able to discover object-material relations as well. Thanks to the ability of our model to predict a label of both object and material at each pixel, it is straightforward to align objects with their associated materials by counting at each pixel what percentage of each material is in every object. Similar to the scene-object relationship, we build a bipartite graph and show its visualization in the left of Figure~\ref{fig:other_relations}. Using this graph we can infer that some sinks are ceramic while others are metallic; different floors have different materials, such as wood, tile, or carpet. Ceiling and wall are painted; the sky is also ``painted", more like a metaphor. However, we can also see that most of the bed is fabric instead of wood, a misalignment due to the actual objects on the bed. Intuitively, the material of a part in an object will be more monotonous. We show the part-material visualization in the middle of Figure~\ref{fig:other_relations}. 

\myparagraph{Material-texture relations.} One type of material may have various kinds of textures. But what is the visual description of a material? We show the visualization of material-texture relations in the right of Figure~\ref{fig:other_relations}. It is worth noting that although there is a lack of pixel-level annotations for texture labels, we can still generate a reasonable relation graph. For example, a carpet can be described as matted, blotchy, stained, crosshatched and grooved. 

In Table~\ref{tab:auto_generation}, we further show some discovered visual knowledge by UPerNet. For scene-object relations, we choose the objects which appear in at least $30\%$ of a scene. For object-material, part-material and material-texture relations, we choose at most top-3 candidates, filter them with a threshold, and normalize their frequencies. We are able to discover the common objects that form each scene, and how much each object or part is made of some material. The visual knowledge extracted and summarized by UPerNet is in consistent with human knowledge. This knowledge base provides rich information across various types of concepts. We hope such knowledge base can shed light on understanding different scenes for future intelligent agents, and ultimately, understanding the real world.

\section{Conclusion}
This work studies the task of Unified Perceptual Parsing, which aims at parsing visual concepts across scene categories, objects, parts, materials and textures from images. A multi-task network and training strategy of handling heterogeneous annotations are developed and benchmarked. We further utilize the trained network to discover visual knowledge among scenes. 

\section*{Acknowledgement}
We would like to show our gratitude to Daniel Karl I. Weidele from MIT-IBM Watson AI Lab for his comments and revision of an earlier version of the manuscript.

\bibliographystyle{splncs}
\bibliography{eccv2018final}

\end{document}